\documentclass{svproc}

\usepackage{url}
\usepackage{helvet}         % selects Helvetica as sans-serif font
\usepackage{courier}        % selects Courier as typewriter font
\usepackage{color}
\usepackage{makeidx}         % allows index generation
\usepackage{graphicx}        % standard LaTeX graphics tool
\usepackage[misc]{ifsym}                            % when including figure files
\usepackage{multicol}        % used for the two-column index
\usepackage[bottom]{footmisc}% places footnotes at page bottom
\usepackage{tabularx}
\usepackage{multirow}
\usepackage{algorithm}
\usepackage{algorithmic}
\usepackage[center]{caption}
\usepackage{longtable}
\usepackage{rotating}
 \usepackage{amsmath}
\usepackage{subfigure}
\usepackage[utf8]{inputenc}

\begin{document}
\mainmatter              % start of a contribution
\title{Decoding Human Emotions: Analyzing Multi-Channel EEG Data using LSTM Networks}
\titlerunning{Decoding Human Emotions from EEG Data using LSTM Networks }  % abbreviated title (for running head)
\author{Shyam K Sateesh\inst{1} \and Sparsh BK\inst{1} \
 \and Uma D.\inst{1} 
}
\authorrunning{Sateesh et al.} % abbreviated author list (for running head)
\institute{%
  \textsuperscript{1}PES University, Bengaluru, India,\\
  \email{shyamksateesh@gmail.com}
}

\maketitle              % typeset the title of the contribution

\begin{abstract}
Emotion recognition from electroencephalogram (EEG) signals is a thriving field, particularly in neuroscience and Human-Computer Interaction (HCI). This study aims to understand and improve the predictive accuracy of emotional state classification through metrics such as valence, arousal, dominance, and likeness by applying a Long Short-Term Memory (LSTM) network to analyze EEG signals. Using a popular dataset of multi-channel EEG recordings known as DEAP, we look towards leveraging LSTM networks' properties to handle temporal dependencies within EEG signal data. This allows for a more comprehensive understanding and classification of emotional parameter states. We obtain accuracies of 89.89\%, 90.33\%, 90.70\%, and 90.54\%  for arousal, valence, dominance, and likeness, respectively, demonstrating significant improvements in emotion recognition model capabilities. This paper elucidates the methodology and architectural specifics of our LSTM model and provides a benchmark analysis with existing papers.
% We would like to encourage you to list your keywords within
% the abstract section using the \keywords{...} command.
\end{abstract}

\section{Introduction}\label{sec:intro}
EEG is defined as the electrical activity of an alternating type recorded from the scalp surface after being picked up by metal electrodes and conductive media \cite{teplan2002}. The unique ability of EEG signals to provide a very descriptive temporal view of brain activity makes it an indispensable tool for understanding complex human emotional states. This capability is especially critical in contexts where the traditional means of emotion assessment are impractical or unfeasible.\\

In recent years, there has been a necessity for understanding and quantifying emotional responses, which has led to advancements in academic research. This has opened new doors for consumer research, mental health, and assistive technologies. The prospect of its ability to assist individuals who would otherwise not be able to express emotions through traditional ways, such as facial expressions, body language, and speech, makes this one of the exciting fields for EEG-based recognition of emotions. These individuals would include, but not be limited to, people with communication disabilities, for example, aphasia; other conditions encompass Autism Spectrum Disorder (ASD) \cite{boutros2015eeg}, among others; and those who have severe physical disabilities from traumatic brain injuries or other progressive diseases, such as Amyotrophic Lateral Sclerosis (ALS) \cite{pirasteh2024eeg}.\\

Statistically, it has been estimated that in the United States alone, approximately 6.6 million people have been diagnosed with some communication disorder \cite{2012ushealth}. From such figures in the global context, it can be estimated that up to 1\% of the world population has some form of autism spectrum disorder. These people's emotional expression and interpretation remain very conventional, usually exhibiting a failing nature. Such failures, thus, stimulate the need to develop standalone technologies that will independently interpret emotional states from physiological data.\\

EEG-based technologies offer a non-invasive, more direct window into the neural underpinnings of emotion. Since it measures electrical activity, the EEG provides a dynamic mapping of activity in the brain, potentially associated with states of emotion without the need for verbal reports or precise physical gestures. This approach is particularly suitable for those whose neurological conditions impair their effective communication.\\

The rise of Long Short-Term Memory (LSTM), a variation of Recurrent Neural Networks (RNNs), has revolutionized this field with its ability to analyze and classify EEG data at unprecedented success rates. LSTMs, in particular, are very strong at modelling the time-dependent features that underlie EEG data. They can capture such underlying patterns temporally, which indicate different emotional states. This advanced machine learning method elevates the prediction performance for emotion classification systems from EEG. It opens an avenue to building real-time responsive systems that can adapt to the emotional feedback of users in different applications.\\

Furthermore, emotion recognition by EEG is possible in healthcare and societal applications. In healthcare, technology could offer better patient care, for instance, since it can interpret pain, discomfort, or emotional distress that patients might be unable to express. One case for this was gauging emotional states in palliative care cancer patients \cite{ramirez2018eeg}. In special education, EEG in non-verbal students could help teachers and caregivers explore the thoughts and emotional states of the students. This could enable tailor-made educational approaches that are more in sync with the mindset of the students \cite{Suhaimi2020}.\\

This study aims to develop more accurate and specific tools for cognitive emotion recognition, particularly for detecting and interpreting emotional states in persons unable to express themselves by traditional means.

\section{Related Work}
The DEAP dataset, detailed by Koelstra et al. \cite{koelstra2012deap}, has been foundational in the field, providing a rich data source for subsequent research. Also, significant correlations were found between the participant ratings and EEG frequencies. The single-trial classification was performed for arousal, valence, and liking scales using features extracted from the EEG, peripheral, and MCA modalities. The results were shown to be significantly better than random classification.
\\\\Alhagry et al. \cite{alhagry2017emotion} proposed LSTM networks for emotion recognition from raw EEG signals. Their study demonstrates that the LSTM model achieves high average accuracies across three emotional dimensions and outperforms traditional emotion recognition techniques, marking a significant advancement in the field.
\\\\Nie et al. \cite{nie2011eeg} explore the relationship between EEG signals and emotional responses while watching movies, focusing on classifying emotions into positive and negative categories. Their application of a Support Vector Machine (SVM) on processed EEG features resulted in an impressive average testing accuracy of 87.53\%, underscoring the potential of EEG-based methods in practical multimedia applications.
\\\\Li et al. \cite{li2022eeg} provide a comprehensive overview of EEG-based emotion recognition, exploring the integration of psychological theories with physiological measurements. They review various machine learning techniques, from conventional models to advanced computational methods, highlighting key advancements and challenges in the field.
\\\\Zheng et al. \cite{zheng2014eeg} develop an innovative approach by integrating deep belief networks with hidden Markov models for EEG-based emotion classification. Their findings indicate that this combined DBN-HMM model achieves higher accuracy than traditional classifiers, highlighting its effectiveness in leveraging spatial and temporal EEG data dimensions.
\\\\Bhagwat et al. \cite{bhagwat2016human} proposed a novel approach for classifying four primary emotions: happy, angry, crying, and sad, which can be visualized as four quadrants. They used Wavelet Transforms (WT) to extract features from raw EEG signals and employed a Hidden Markov Model (HMM) to classify emotions.
\\\\Lin et al. \cite{lin2010eeg} utilize EEG data and machine learning to enhance emotional state predictions during music listening. Using an SVM, their approach achieves an average classification accuracy of 82.29\% for emotions such as joy, anger, sadness, and pleasure. 
\\\\Naser and Saha \cite{naser2013recognition} applied advanced signal processing techniques to improve feature extraction for emotion classification from EEG signals. Their study utilizes dual-tree complex wavelet packet transform (DT-CWPT) and statistical methods like QR factorization and singular value decomposition (SVD) to select discriminative features effectively. The enhanced feature set is then classified using an SVM, demonstrating notable improvements in classification accuracy.
\\\\Li et al. \cite{li2017channel} found that while it is feasible to work with single-channel EEG data, it is much more effective to combine multiple channels of EEG features into a single feature vector. They also found that the beta and gamma frequency bands are more related to emotional processing than the other bands.
\\\\The exploration of adaptive emotion detection using EEG and the Valence-Arousal-Dominance model by Gannouni et al. \cite{gannouni2020adaptive} advances the field by adapting computational models to individual brain activity variations. Their method employs an adaptive selection of electrodes, significantly enhancing emotion detection accuracy. Utilizing machine learning algorithms, the study demonstrates a 5\% and 2\% increase in accuracy for valence, arousal, and dominance dimensions, respectively, compared to fixed-electrode approaches.
\\\\Alvarez-Jiménez et al. \cite{alvarez2024comprehensive} enhance EEG-based emotion recognition by integrating diverse feature sets from multiple domains. Their use of various classifiers, including Artificial Neural Networks, achieves a high accuracy of 96\%, demonstrating the effectiveness of hybrid features in improving model robustness.
\\\\Atkinson-Abutridy et al. \cite{atkinson2015improving} proposed a feature-based emotion recognition model combining statistical-based feature selection methods with SVM classifiers, focusing on Valence/Arousal dimensions for classification. This combined approach outperformed other recognition methods. 
\\\\Yoon and Chung \cite{yoon2013eeg} detailed a probabilistic classifier based on Bayes' theorem and a supervised learning approach using a perceptron convergence algorithm, offering a methodologically distinct perspective on emotion classification from EEG signals.

\section{\textbf{Dataset}}

The Database for Emotion Analysis using Physiological Signals (DEAP) \cite{koelstra2012deap} is at the core of our study, and it presents a rich source of EEG and peripheral physiological signals for analyzing emotions. The dataset was built to boost and proliferate the development of systems that would be capable of recognizing human emotions from physiological responses, with particular emphasis on the paradigms of human-computer interaction.

\subsection{\textbf{Dataset Description }}
The DEAP dataset consists of EEG data recordings from 32 participants between 19 and 37 years old, with a mean age of 26.9 years. Each participant was presented with 40 one-minute music video clips to elicit emotional responses. Participants had to rate their experience after each stimulus on a 1 to 9 integer scale for arousal, valence, dominance, and liking. We will use these subjective ratings as labels to train our models.

\subsection{\textbf{Data Acquisition }}

EEG and peripheral physiological signals were acquired simultaneously, viewing each music video clip by all the participants. In the course of the experiment, the recording of the EEG data was carried out at 512 Hz through the 32-channel systems, which was eventually reduced to 128 Hz during analysis. Concurrently with EEG, other physiological signals such as galvanic skin response and heart rate were also recorded to deliver complete states regarding the participant's physiological states during each trial.

\subsection{\textbf{Data Structure }}

For each participant, it is composed of two main arrays: the EEG signals and an array of labels for each trial. The EEG data array has a dimension of 40x40x8064 for 40 trials, 40 channels, and 8064 data points per channel per trial. Corresponding to four emotional dimensions assessed per video clip, the array structure of labels is 40x4. (Shown in Table 1)

\begin{table}[htbp]
    \centering
    \caption{DEAP: Structure of each participant array}
    \label{tab:deap_structure}
    \renewcommand{\arraystretch}{1.5} % Adjust the value as needed for more vertical space
    \begin{tabular}{|l|c|l|}
        \hline
        \textbf{Array Name} & \textbf{Array Shape} & \textbf{Array Contents} \\
        \hline
        \hline
        data & 40 x 40 x 8064 & video/trial x channels x data \\
        \hline
        labels & 40 x 4 & video/trial x label (VADL) \\
        \hline
    \end{tabular}
\end{table}

\section{\textbf{Valence, Arousal, and Dominance Model}}
\vspace{\baselineskip}
The Valence-Arousal-Dominance (VAD) model presents a sensitive framework for recognizing human emotions and classifies them into three significant aspects: valence, arousal, and dominance. Valence measures the 'how good or bad' of the mood, arousal measures the activation level, and dominance measures how much control one might feel they have over their emotional state.
Researchers have adopted this model in examining features of EEG that indicate the functioning of different areas in the brain toward emotional stimuli. Research has shown that the positive effect increases alpha band activity in the frontal regions. In contrast, the negative one tends to decrease it, and high arousal corresponds to beta activity.
\\\\The usage of recent advanced classification techniques, even with segmenting EEG data into minor epochs, has increased the accuracy of emotional assessments. Such improvements enhance the accuracy by around 5\% for valence and arousal and 2\% for dominance, respectively, hence depicting the effectiveness of the VAD model in the subtle multi-dimensionality characterization of human emotions \cite{gannouni2020adaptive}. (VAD Model Shown in Fig. 1 \cite{gannouni2020adaptive})

\begin{figure}
    \centering
    \includegraphics[width=0.65\linewidth]{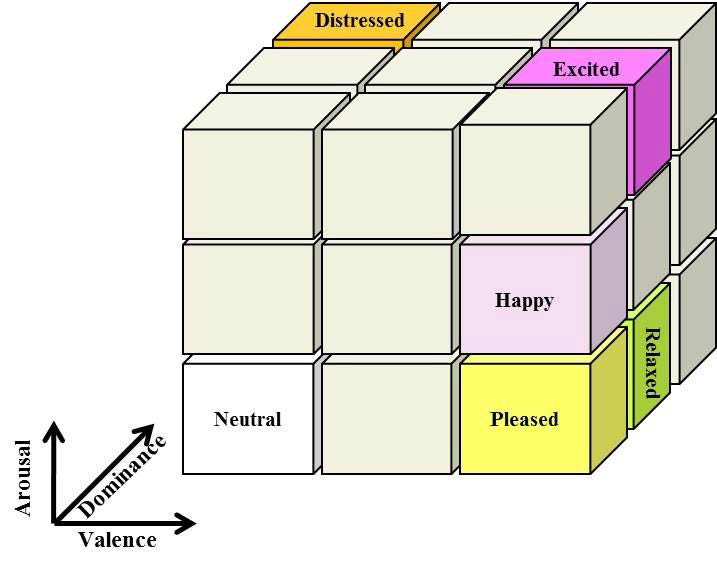}
    \caption{Valence-Arousal-Dominance Model Depiction as A 3-D Graph \cite{gannouni2020adaptive} }  
\end{figure}

\section{\textbf{Rationale Behind Usage of Deep Learning}}

Emotion recognition from EEG data is a difficult task due to the complexity and variability of the signals. Although traditional statistical methods effectively analyze structured and more straightforward datasets, they frequently fail to capture and interpret the dynamic and non-linear interactions typical of EEG data. Deep learning, a branch of machine learning, has become an invaluable tool for managing these complexities due to its ability to discern high-level, abstract features from vast amounts of data.

\subsection{\textbf{Deep Learning vs. Traditional Statistical Methods}}
Deep learning models that handle unstructured data, like images, speech, or biological signals, perform this function due to their use of neural networks. Traditional statistical approaches to data analysis require manual choice of features, and at most, they can only model the linear effects. This is essential in EEG data, where emotional states are not explicitly encoded but latent constructs reflected in slight signal variations. Deep learning models enable learning such patterns directly from raw data, optimizing feature extraction, selection, and classification tasks in a joint form. This shows that robust and accurate analysis is developed in high dimensionality and noise levels that are usually related to EEG recordings.

\subsection{\textbf{Overview of LSTMs}}
LSTMs are one of the unique variants of Recurrent Neural Networks (RNN). It was first introduced by Hochreiter et al. \cite{hochreiter1997long} to eliminate the problem of long-term dependencies seen in conventional RNNs. Traditional RNNs are known to also suffer from gradient-related issues. This problem, in turn, makes it very hard for them to be trained on sequential data where long-term contextual information is essential. LSTMs solve this problem due to the exceptional structure of their gates, which allows them to regulate the flow of information in a way that enables them to remember or forget information for long periods.

\subsection{\textbf{Bidirectional LSTMs}}
The capabilities of standard LSTMs are further advanced through the usage of bidirectional LSTMs, allowing more context to be available from the subsequent points in the data sequence. So, bidirectional LSTMs can capture the context information from past and future states by processing data in the forward and reverse directions. This is very useful in emotion recognition from the EEG signals when the emotional state reflected in a data segment may depend not only on the earlier but also on the latter events.

\subsection{\textbf{LSTM for Our Work}}
In this project, we choose LSTM networks due to their prowess in sequence prediction problems, thus capable of adequately modelling the temporal dynamics characteristic of EEG data. Applying LSTMs will help reach the deepest emotional timelines that fall within the EEG signals, making them more helpful in predicting emotional states with better accuracy. The bidirectional approach of the capability enforces the complete context from all the data points, which increases the recognition accuracy for complex emotional states. This makes LSTMs very apt for the development of a robust system for emotion recognition from EEG-based data.

\section{\textbf{Proposed Method} }
\vspace{\baselineskip}
Our study uses an LSTM model to classify emotional states from EEG data and focuses on feature extraction, data preparation, and architectural considerations to achieve high accuracy percentages.

\begin{figure}
    \centering
    \includegraphics[width=0.9\linewidth]{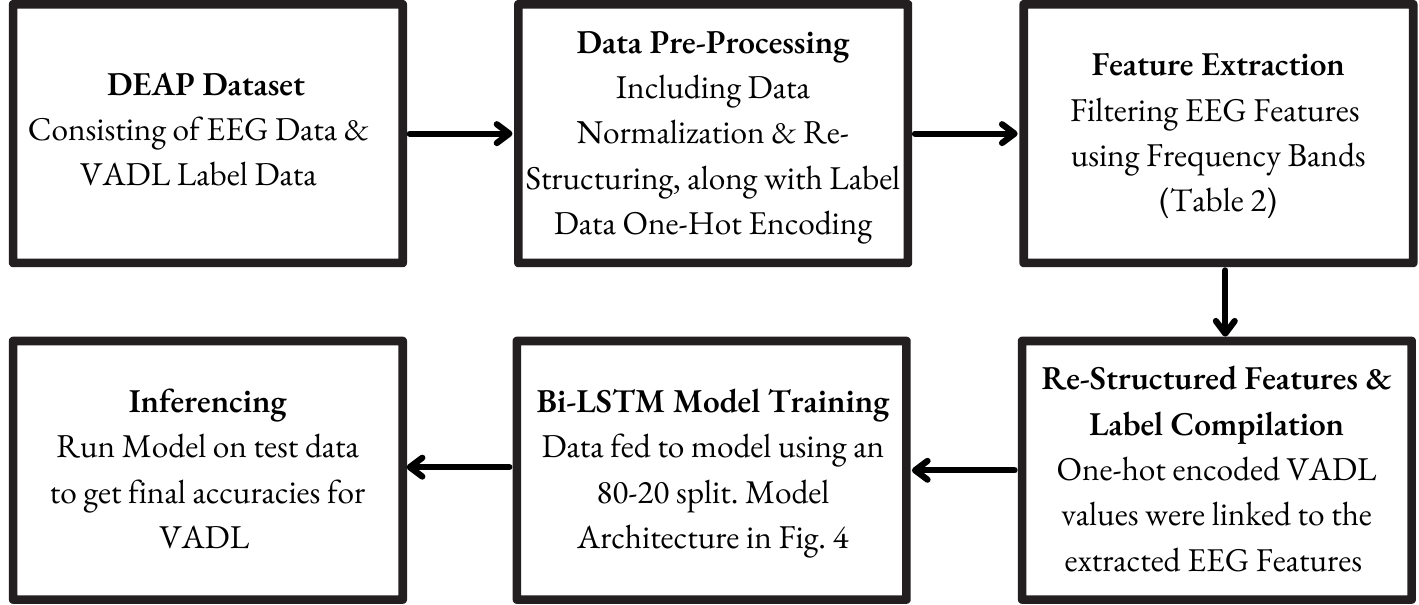}
    \caption{Flowchart of Proposed Scheme}
    
\end{figure}

\subsection{\textbf{Pre-Processing Methods Used }}
The feature extraction process was tailored to capture significant information from the EEG signals. We utilized specific EEG channels and frequency bands relevant to emotional processing. The chosen channels included a subset correlating with emotional states, such as frontal and temporal regions. Frequency bands were segmented into five distinct ranges: theta (4-8 Hz), alpha (8-12 Hz), low beta (12-16 Hz), high beta (16-30 Hz), and gamma (30-45 Hz), which are traditionally associated with different aspects of cognitive processing and emotional regulation (Refer to Table 2 and Fig. 3 \cite{featurediagram}). Each of these bands aids in extracting vital information from input EEG data, which has been proven to support sentiment analysis \cite{featurediagram}. The Fast Fourier Transform (FFT) process was applied to a select 14 channels of the recorded 32 channels, chosen to fit Emotiv Epoc, with a window size of 256 points, corresponding to 2 seconds of data, with an overlap of 0.125 seconds to ensure comprehensive temporal analysis. 

\begin{table}[htbp]
    \centering
    \caption{EEG Feature Bands Used in the Study}
    \label{tab:frequency_bands}
    \renewcommand{\arraystretch}{1.5} % Adjust the value as needed for more vertical space
    \resizebox{0.9\textwidth}{!}{%
        \begin{tabular}{|c|c|p{5cm}|}
            \hline
            \textbf{Brainwave Type} & \textbf{Frequency Range (Hz)} & \textbf{Mental States \& Conditions Seen} \\
            \hline
            \hline
            Theta & 4 - 8 & Intuitive, creative, recall, fantasy, imaginary, dream \\
            \hline
            Alpha & 8 - 12 & Relaxed but not drowsy, calm, conscious \\
            \hline
            Low-Beta & 12 - 16 & Relaxed yet focused, integrated \\
            \hline
            High-Beta & 16 - 30 & Alertness, agitation \\
            \hline
            Gamma & 30 - 45 & Cognition, information processing \\
            \hline
        \end{tabular}%
    }
\end{table}

\begin{figure}
    \centering
    \includegraphics[width=0.7\linewidth]{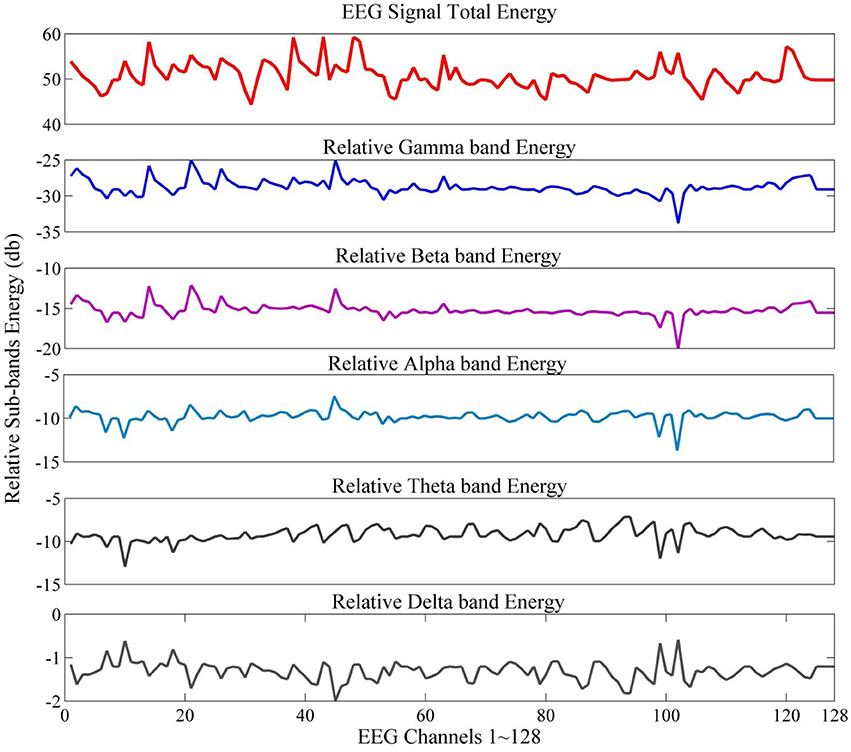}
    \caption{EEG signal energy and relative sub-band energy \cite{featurediagram}}
    
\end{figure}

%\subsection{\textbf{Additional Pre-Processing Used }}

The dataset was first split into training and test splits using an 80-20 ratio. That is, 80\% of the data was used to train the LSTM model, and the remaining 20\% was utilized to test the model's performance. This split ratio enabled the practical training of the model as well as a reliable evaluation to establish how it generalizes to unseen data.
 
%\subsection{\textbf{Normalization }}

Data Normalization was necessary to normalize the input features to reduce discrepancies in signal amplitudes caused by data variations across individuals. All feature vectors were normalized to zero mean and unit variance, a standard approach in processing EEG signals to overcome inter-subject differences.

%\subsection{\textbf{Conversions to One-Hot Encodings }}

 The next step was converting each valence, arousal, dominance, and likeness label (initially scaled from 1-9) into one-hot encodings to create nine classes before sending the label data into the LSTM Networks. This was implemented using the $keras.utils.to\_categorical()$ function.

\subsection{\textbf{LSTM Architecture }}

The LSTM network architecture employed in our study is designed to handle EEG data sequentially and temporally effectively. We will use one LSTM model for each emotional parameter in observation. The model initiates with a Bidirectional LSTM layer consisting of 128 units, enhancing the model's ability to capture dependencies in both forward and backward directions of the input sequence. This layer is followed by a dropout of 0.6 to reduce overfitting by randomly ignoring a fraction of the neurons during training.\\

Subsequent layers include multiple LSTM layers with varying numbers of neurons to extract and refine features from the data incrementally. Specifically, the model includes an LSTM layer with 256 units and another two LSTM layers, each with 64 units, all incorporating a dropout of 0.6 after each LSTM layer to prevent overfitting further. The final LSTM layer consists of 32 units, followed by a dropout of 0.4, aiming to consolidate the features extracted by previous layers into a more manageable form.\\

The output from the LSTM layers is then passed through two dense layers. The first dense layer has 16 units with a ReLU activation function intended to introduce non-linearity into the model, facilitating the network's ability to learn complex patterns. The final output layer consists of some units equal to the classes of emotions being classified, with a softmax activation function to output the probability distribution over the classes.\\

This architecture is compiled with the Adam optimizer and categorical cross-entropy as the loss function, suitable for multi-class classification problems. The detailed structure and parameterization of the model are crucial for its ability to discern nuanced emotional states from EEG data, as visualized in the accompanying architectural diagram in our study. A representation of the model is shown in Fig. 4.

\begin{figure}
    \centering
    \includegraphics[width=0.65\linewidth]{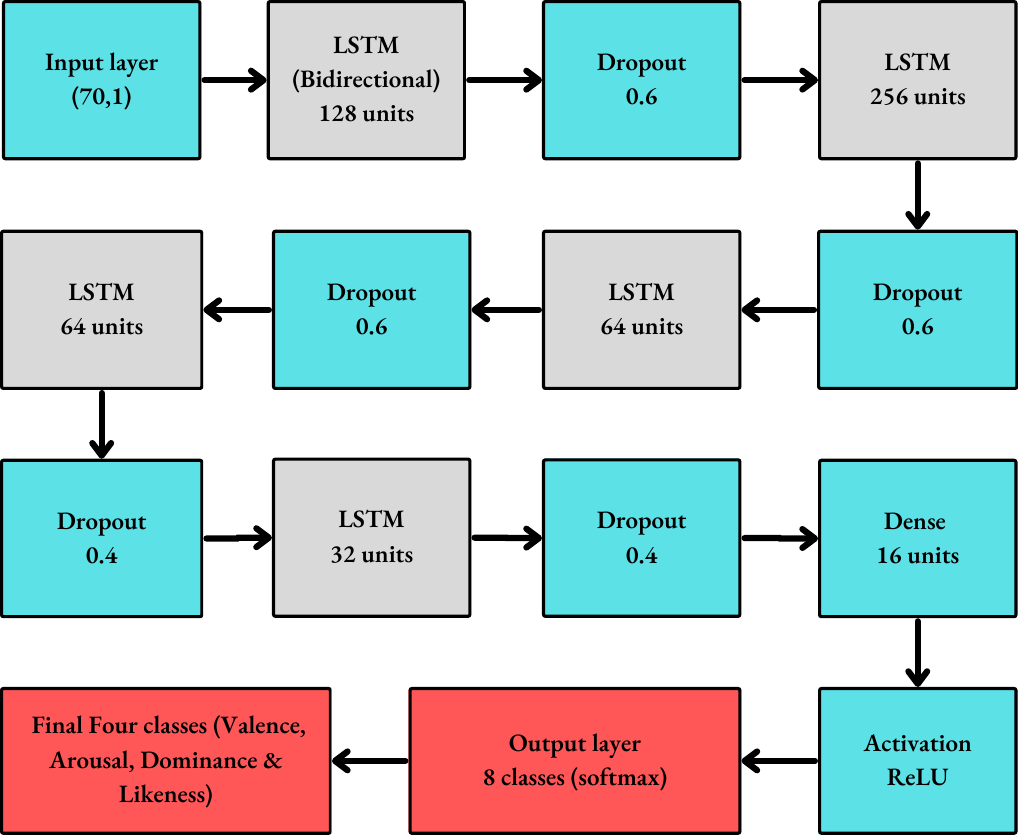}
    \caption{Our LSTM Architecture}
    
\end{figure}

\section{  \textbf{Results} }
\vspace{\baselineskip}
Our LSTM-based model demonstrated outstanding performance in emotion recognition from EEG data, achieving individual class accuracies of 90.33\% for valence, 89.89\% for arousal, 90.70\% for dominance, and 90.54\% for likeness, with an overall accuracy of 90.36\%. These results underline the model's efficacy in capturing complex emotional states through advanced feature extraction and a robust LSTM architecture. This performance showcases the model's capabilities and sets a foundation for future advancements in EEG-based emotion recognition. A comparison of accuracies is attached in Table 3, showing our method of using an LSTM network to be highly accurate and effective in classifying emotional parameters correctly compared to related papers.

\begin{table}[htbp]
    \centering
    \caption{Average Accuracies and Nature of Features Extracted}
    \label{tab:average_accuracy}
    \renewcommand{\arraystretch}{1.5} % Adjust the value as needed for more vertical space
    \resizebox{0.86\textwidth}{!}{%
        \begin{tabular}{|l|c|c|c|c|c|}
            \hline
            \textbf{Paper} & \textbf{Arousal} & \textbf{Valence} & \textbf{Liking} & \textbf{Features Extracted} & \textbf{Frequency Bands Ranges (Hz)} \\
            \hline
            \hline
            Koelstra et al. \cite{koelstra2012deap} & 62.00 & 56.70 & 55.40 & Frequency Based & (4-7), (8-13), (14-29), (30-47) \\
            Atkinson and Campos \cite{atkinson2015improving} & 73.06 & 73.41 & -- & Statistical Based & - \\
            Yoon and Chung \cite{yoon2013eeg} & 70.10 & 70.90 & -- & Frequency Based & (4-8), (8-13), (13-30), (36-44) \\
            Naser and Saha \cite{naser2013recognition} & 66.20 & 64.30 & 70.20 & Time-Frequency Based & - \\
            Alhagry et al. \cite{alhagry2017emotion} & 85.65 & 85.45 & 87.99 & Frequency Based & (4-8), (8-10), (8-12), (12-30), (30+) \\
            Li et al. \cite{li2022eeg} & 83.78 & 80.72 & -- & Frequency Based & (4-8), (8-13), (13-30), (30-45) \\
            Acharya D et al. \cite{acharyaeeg}& -- & -- & 88.60 & Frequency Based & (4-8), (8-12), (12-16), (16-25), (25-45) \\
            \hline
            \textbf{Proposed Method} & \textbf{89.89} & \textbf{90.33} & \textbf{90.54} & \textbf{Frequency Based} & \textbf{(4-8), (8-12), (12-16), (16-30), (30-45)} \\
            \hline
        \end{tabular}%
    }
\end{table}

\section{\textbf{Conclusion} }
\vspace{\baselineskip}
This study successfully showcases the efficacy of LSTM networks in accurately classifying emotional states from EEG data, achieving high performance across various emotional dimensions. The customized LSTM architecture, incorporating bidirectional layers and strategic dropout stages, adeptly handles the complexities of EEG signals. Our LSTM architecture paired with uniform frequency band ranges taken for EEG feature extraction has proven to provide improved results from previous LSTM-based EEG studies. Such capabilities pave the way for advancements in cognitive neuroscience and human-computer interaction, promising enhancements in responsive systems that adapt to user emotions in real time. Future work can further build upon this model with more robust neural networks, including time-frequency and location domain features, along with the possible usage of more than 14 EEG channels for better efficiency of emotion recognition. Upcoming research will benefit from exploring hybrid models that integrate additional physiological signals, further refining the precision and application of EEG-based emotion recognition in creating empathetic user interfaces.

\bibliographystyle{plain}

\begin{thebibliography}{10}

\bibitem{acharyaeeg}
Divya Acharya, Riddhi Jain, Siba~Smarak Panigrahi, Rahul Sahni, Siddhi Jain, Sanika~Prashant Deshmukh, and Arpit Bhardwaj.
\newblock Multi-class emotion classification using eeg signals.
\newblock In Deepak Garg, Kit Wong, Jagannathan Sarangapani, and Suneet~Kumar Gupta, editors, {\em Advanced Computing}, pages 474--491, Singapore, 2021. Springer Singapore.

\bibitem{alhagry2017emotion}
Salma Alhagry, Aly~Aly Fahmy, and Reda~A. El-Khoribi.
\newblock Emotion recognition based on eeg using lstm recurrent neural network.
\newblock {\em International Journal of Advanced Computer Science and Applications (IJACSA)}, 8(10), 2017.

\bibitem{pirasteh2024eeg}
Manouchehr Shamseini~Ghiyasvand Alireza~Pirasteh and Majid Pouladian.
\newblock Eeg-based brain-computer interface methods with the aim of rehabilitating advanced stage als patients.
\newblock {\em Disability and Rehabilitation: Assistive Technology}, 0(0):1--11, 2024.
\newblock PMID: 38400897.

\bibitem{featurediagram}
Hafeez~Ullah Amin, Wajid Mumtaz, Ahmad~Rauf Subhani, Mohamad Naufal~Mohamad Saad, and Aamir~Saeed Malik.
\newblock Classification of eeg signals based on pattern recognition approach.
\newblock {\em Frontiers in Computational Neuroscience}, 11, 2017.

\bibitem{atkinson2015improving}
John Atkinson and Daniel Campos.
\newblock Improving bci-based emotion recognition by combining eeg feature selection and kernel classifiers.
\newblock {\em Expert Systems with Applications}, 47:35--41, 2016.

\bibitem{bhagwat2016human}
Anuja~R. Bhagwat and A.~N. Paithane.
\newblock Human disposition detection using eeg signals.
\newblock In {\em 2016 International Conference on Computing, Analytics and Security Trends (CAST)}, pages 366--370, 2016.

\bibitem{boutros2015eeg}
Nash~N. Boutros, Renee Lajiness-O'Neill, Andrew Zillgitt, Anette~E. Richard, and Susan~M. Bowyer.
\newblock Eeg changes associated with autistic spectrum disorders.
\newblock {\em Neuropsychiatric Electrophysiology}, 1(1):3, 2015.

\bibitem{gannouni2020adaptive}
Sofien Gannouni, Arwa Aledaily, Kais Belwafi, and Hatim Aboalsamh.
\newblock Adaptive emotion detection using the valence-arousal-dominance model and eeg brain rhythmic activity changes in relevant brain lobes.
\newblock {\em IEEE Access}, 8:67444--67455, 2020.

\bibitem{hochreiter1997long}
Sepp Hochreiter and Jürgen Schmidhuber.
\newblock {Long Short-Term Memory}.
\newblock {\em Neural Computation}, 9(8):1735--1780, 11 1997.

\bibitem{koelstra2012deap}
Sander Koelstra, Christian Muhl, Mohammad Soleymani, Jong-Seok Lee, Ashkan Yazdani, Touradj Ebrahimi, Thierry Pun, Anton Nijholt, and Ioannis Patras.
\newblock Deap: A database for emotion analysis ;using physiological signals.
\newblock {\em IEEE Transactions on Affective Computing}, 3(1):18--31, 2012.

\bibitem{li2022eeg}
Xiang Li, Yazhou Zhang, Prayag Tiwari, Dawei Song, Bin Hu, Meihong Yang, Zhigang Zhao, Neeraj Kumar, and Pekka Marttinen.
\newblock Eeg based emotion recognition: A tutorial and review.
\newblock {\em ACM Comput. Surv.}, 55(4), nov 2022.

\bibitem{li2017channel}
{Li, Xian}, {Yan, Jian-Zhuo}, and {Chen, Jian-Hui}.
\newblock Channel division based multiple classifiers fusion for emotion recognition using eeg signals.
\newblock {\em ITM Web Conf.}, 11:07006, 2017.

\bibitem{lin2010eeg}
Yuan-Pin Lin, Chi-Hong Wang, Tien-Lin Wu, Shyh-Kang Jeng, and Jyh-Horng Chen.
\newblock Eeg-based emotion recognition in music listening: A comparison of schemes for multiclass support vector machine.
\newblock In {\em 2009 IEEE International Conference on Acoustics, Speech and Signal Processing}, pages 489--492, 2009.

\bibitem{2012ushealth}
Megan~A. Morris, Sarah~K. Meier, Joan~M. Griffin, Megan~E. Branda, and Sean~M. Phelan.
\newblock Prevalence and etiologies of adult communication disabilities in the united states: Results from the 2012 national health interview survey.
\newblock {\em Disability and Health Journal}, 9(1):140--144, 2016.

\bibitem{naser2013recognition}
Daimi~Syed Naser and Goutam Saha.
\newblock Recognition of emotions induced by music videos using dt-cwpt.
\newblock In {\em 2013 Indian Conference on Medical Informatics and Telemedicine (ICMIT)}, pages 53--57, 2013.

\bibitem{nie2011eeg}
Dan Nie, Xiao-Wei Wang, Li-Chen Shi, and Bao-Liang Lu.
\newblock Eeg-based emotion recognition during watching movies.
\newblock In {\em 2011 5th International IEEE/EMBS Conference on Neural Engineering}, pages 667--670, 2011.

\bibitem{ramirez2018eeg}
Rafael Ramirez, Josep Planas, Nuria Escude, Jordi Mercade, and Cristina Farriols.
\newblock Eeg-based analysis of the emotional effect of music therapy on palliative care cancer patients.
\newblock {\em Frontiers in Psychology}, 9, 2018.

\bibitem{Suhaimi2020}
Nazmi~Sofian Suhaimi, James Mountstephens, and Jason Teo.
\newblock Eeg-based emotion recognition: A state-of-the-art review of current trends and opportunities.
\newblock {\em Computational Intelligence and Neuroscience}, 2020(1):8875426, 2020.

\bibitem{teplan2002}
Michal Teplan.
\newblock Fundamental of eeg measurement.
\newblock {\em MEASUREMENT SCIENCE REVIEW}, 2, 01 2002.

\bibitem{yoon2013eeg}
Hyun~Joong Yoon and Seong~Youb Chung.
\newblock Eeg-based emotion estimation using bayesian weighted-log-posterior function and perceptron convergence algorithm.
\newblock {\em Computers in Biology and Medicine}, 43(12):2230--2237, 2013.

\bibitem{zheng2014eeg}
Wei-Long Zheng, Jia-Yi Zhu, Yong Peng, and Bao-Liang Lu.
\newblock Eeg-based emotion classification using deep belief networks.
\newblock volume 2014, 07 2014.

\bibitem{alvarez2024comprehensive}
Mayra Álvarez Jiménez, Tania Calle-Jimenez, and Myriam Alvarez.
\newblock A comprehensive evaluation of features and simple machine learning algorithms for electroencephalographic-based emotion recognition.
\newblock {\em Applied Sciences}, 14:2228, 03 2024.

\end{thebibliography}
\fontsize{10}{10.6}\selectfont
%\addcontentsline{toc}{section}{\textbf{REFERENCES}}
%\renewcommand{\bibname}{REFERENCES}

%\end{spacing}

\end{document}